# Instant-LIVO: Efficient LiDAR-Inertial-Visual Odometry with Compact Geometry and a Bounded Single-Reference-Frame Map

Xinrui Wang[1], Ziwei Lan[1,2,3], Wenbin He[1,2,3], Mingjing Yang[1,2,3], Yihan Zhang[1], Yilin Liu[1], Peiji Chen[1], Wenxin Li[1], Shuyi Jiang[1], Yubin Lan[1,2,3],Weiguang Yang[1,2,3,*]

*Abstract*— **LiDAR-Inertial-Visual Odometry (LIVO) is a fundamental technology for autonomous systems; however, deployment on resource-constrained platforms remains challenging due to physical limitations. This paper presents Instant-LIVO, a tightly coupled, high-speed, real-time LIVO system. The system introduces a compact geometric map based on spatial-hashed OctVox and accelerates local-neighborhood retrieval and plane fitting through cache-assisted heuristic staged K-nearest-neighbor searches. It also employs a bounded single-reference-frame visual map that adaptively updates the reference frame according to the tracking status, while using Schmidt-type uncertainty-aware photometric fusion to maintain the correlation between the reference pose and the current state. The system performs multisensor fusion through a sequential process consisting of IMU prediction, LiDAR geometric correction, and visual photometric correction. It was evaluated on 32 public benchmarks and custom datasets. While maintaining competitive accuracy on both X86 and ARM platforms, Instant-LIVO achieves average processing-speed improvements of approximately 2.3× and 2.1×, respectively, and reduces peak memory usage to approximately 1/20 and 1/13, respectively. The results demonstrate that the proposed method outperforms existing LIO and LIVO baselines in both accuracy and computational efficiency. The implementation is available at: https://github.com/Faker-0714/Instant-LIVO.git.**

## I. Introduction

Reliable motion estimation and environment reconstruction are fundamental to the operation of autonomous robots in unknown environments. Vision-based methods, such as ORB-SLAM3 [1], estimate motion from image texture features; however, their visual constraints are limited by imaging conditions and become ineffective under low-light or no-light conditions. LiDAR-based methods, such as LOAM [2] , provide direct geometric measurements, but their motion constraints may weaken in environments with insufficient geometric variation. Inertial measurements support high-frequency motion prediction, but integration errors accumulate over time. These complementary sensing characteristics have motivated researchers to fuse camera, LiDAR, and inertial measurements to achieve robust state estimation.

*Corresponding author: Weiguang Yang (wgyang@scau.edu.cn)

[1]College of Artificial Intelligence and Low-Altitude Technology, South China Agricultural University, Guangzhou 510642, China

[2]National Center for International Collaboration Research on Precision Agricultural Aviation Pesticides Spraying Technology (NPAAC), South China Agricultural University, Guangzhou 510642, China

[3]Low-Altitude Research Institute, South China Agricultural University, Guangzhou 510642, China

In recent years, progress has been made in both geometric estimation and multisensor fusion. FAST-LIO2 [3] combines direct point-cloud registration with an incremental k-d tree, whereas Faster-LIO [4] accelerates spatial queries using sparse incremental voxels. For further integration of visual measurements, LVI-SAM [5] couples visual–inertial and LiDAR–inertial subsystems within a factor-graph framework, while R3LIVE [6] introduces photometric observations for state estimation and color mapping. FAST-LIVO [7] associates image patches with LiDAR map points to enable sparse direct registration. FAST-LIVO2 [8] further integrates sequential filtering, plane-assisted image-patch warping, and online exposure estimation within a direct LiDAR–inertial–visual odometry (LIVO) framework.

Efficient map access and visual-data management provide significant advantages for deploying LIVO on computationally resource-constrained platforms. Zhou et al. [9] improved performance through adaptive visual-frame selection and a map structure that combines local geometric and visual information with long-term visual-data storage. However, research on improving efficiency without compromising accuracy remains limited.

Several key technical challenges must be addressed to achieve these advantages. First, simplifying geometric information may result in the loss of local surface details. Therefore, the geometric structures required for point-to-plane registration and image-patch warping must be preserved while reducing map-storage and query overhead. Second, limited visual-data storage constrains the available historical observations. It is therefore necessary to maintain continuous and reliable photometric constraints under viewpoint changes and to properly handle uncertainty and correlation during successive visual updates, thereby balancing computational efficiency with state-estimation consistency.

To address these challenges, we propose Instant-LIVO, a tightly coupled LIVO system built upon FAST-LIVO2 [8]. The system combines a compact OctVox geometric map with a bounded-size single-reference-frame visual representation.

The main contributions of this paper are as follows:

1) **Compact OctVox [16] mapping**: Hash voxels retain up to eight representative child-subvoxel points per block. Cache-accelerated C-HKNN performs staged searches with distance-bound early termination; on-demand planes support point-to-plane registration and image-patch warping.

2) **Single-reference-frame map and uncertainty-aware photometric fusion**: A bounded reference-frame map is adaptively replaced according to tracking status. Schmidt photometric updates [10] preserve reference-pose uncertainty and state correlation through IMU prediction, LiDAR correction, and visual fusion.

3) **Extensive experimental evaluation**: The tightly coupled LIVO framework is comprehensively evaluated on public benchmark datasets and self-collected datasets.

## II. Related Works

### *A. LiDAR-Visual-Inertial Integration Evolution*

LiDAR–visual–inertial fusion addresses the limitation of relying on a single type of constraint in pure LIO. R2LIVE [11] uses sparse visual features and minimizes reprojection errors. FAST-LIVO2 [8] further improves this approach through sequential ESIKF updates and a direct method. GS-LIVO [12] introduces 3D Gaussian Splatting into a real-time LIVO framework. LIR-LIVO [13] incorporates SuperPoint and LightGlue to enhance visual-feature extraction and matching. Voxel-LIVO [14] constructs a unified adaptive voxel map for efficient data association. Existing studies have made progress in fusion accuracy, robustness, and map representation; however, computational efficiency under resource-constrained conditions still has room for improvement.

### *B. Geometric Map Design*

The organization of a geometric map directly affects the computational cost of point-cloud maintenance, neighborhood retrieval, and geometric-residual construction. LOAM [2] adopts a feature-based map representation and extracts point clouds into edge and planar features for storage. LIO-SAM [15] constructs a local map by selecting a fixed number of historical sub-keyframes. FAST-LIO2 [3] employs an incremental k-d tree (ikd-Tree) to achieve efficient nearest-neighbor searches. Faster-LIO [4] uses a sparse incremental voxel structure (iVox) to reduce query overhead. FAST-LIVO2 [8] extends the map into a unified visual–geometric voxel structure and associates image patches with the LiDAR geometric map. Super-LIO [16] reduces redundant nearest-neighbor retrieval computations by using an octree voxel map (OctVox) and a heuristic nearest-neighbor search strategy (HKNN). Mature map-organization and management methods in the LIO field provide useful guidance for map design in LIVO systems.

### *C. Visual Data Management*

The core objective of visual-data management is to maintain usable photometric constraints from a limited set of historical observations. LVI-SAM [5] and R3LIVE [6] maintain visual information across time for smoothing-based mapping and color mapping, respectively. FAST-LIVO [7] and FAST-LIVO2 [8] reduce the overhead of feature description and matching through sparse image patches and direct photometric residuals. However, existing methods generally rely on multi-frame historical images or continuously maintained visual maps, and therefore require the management of a relatively large amount of historical observation information.

### *D. Filtering Approaches*

The filtering approach determines the order in which multisensor observations are fused during recursive estimation and whether historical states and map information continue to participate in subsequent updates. LVI-SAM [5] employs factor-graph smoothing for cross-modal state optimization, whereas FAST-LIVO2 [8] organizes IMU propagation, LiDAR geometric updates, and visual updates into a sequential filtering process. FAST-LIO2 [3] uses an iterated error-state Kalman filter to process direct geometric residuals. For long-term visual-map maintenance, SEVIS [10] converts mature map features into Schmidt states. While keeping their state means fixed, it preserves their cross-covariances with the current active states, thereby reducing the computational overhead caused by the participation of historical map features in filter updates.

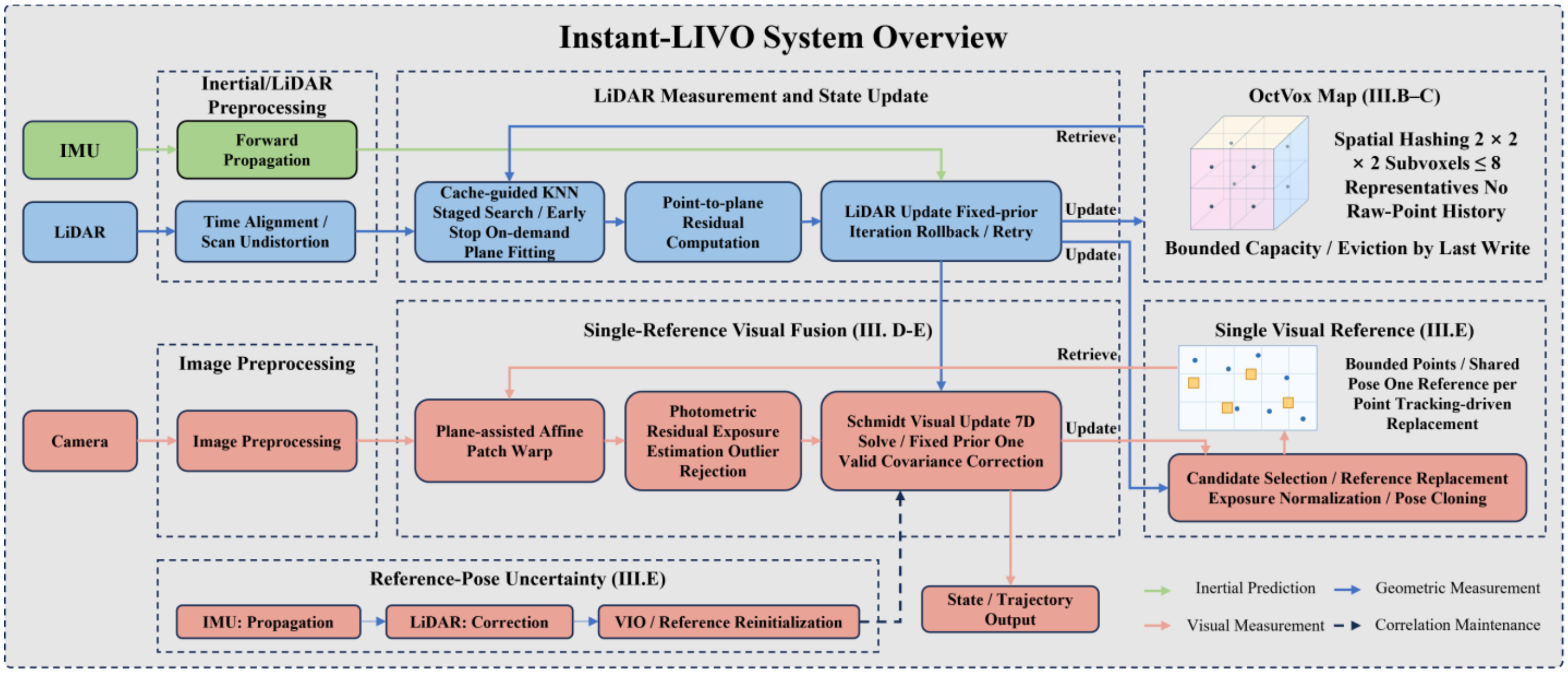

Figure 1. Overview of the Instant-LIVO system. The green, blue, and red modules represent IMU, LiDAR, and visual processing, respectively; arrows of different colors denote inertial prediction, geometric measurements, visual measurements, and correlation maintenance.

## III. System Overview And Methodology

### A. System Overview

Instant-LIVO adopts a sequential processing pipeline consisting of IMU prediction, LiDAR correction, and visual correction, as illustrated in Fig. 1. OctVox provides local geometric information through a compact representation and cached staged retrieval, while the single-reference-frame visual map provides bounded photometric observations and handles reference-pose uncertainty through Schmidt-type updates. The system synchronizes the sensor data at the image timestamps, propagates the state using IMU measurements, and compensates for motion distortion in the point cloud. The state after LIO processing serves as the prior for the visual update; when the LiDAR geometric correction is rejected, the visual update instead uses the IMU prediction. The aforementioned time synchronization, inertial processing, and basic iterative filtering follow FAST-LIVO2 [8].

### B. Octo-Voxel-Based Map (OctVox)

OctVox organizes voxel blocks using spatial hashing and divides each block into eight subvoxels. Each subvoxel uses the sample mean as its geometric representative, which is recursively updated subject to sample-count and spatial-distance constraints. Let $\mu_s$ and $n_s$ denote the representative point and the number of accepted samples, respectively, for subvoxel $s$, and let $p_i^W$ denote the world coordinates of the current new sample; the mean is then updated recursively as

$$\boldsymbol{\mu}_s^+ = \boldsymbol{\mu}_s + \frac{\mathbf{p}_i^W - \boldsymbol{\mu}_s}{n_s + 1}, n_s^+ = n_s + 1 \quad (1)$$

The first sample is used to initialize the representative point. Local planes are fitted on demand using representative points in the neighborhood of the query point to construct geometric constraints.

The map is updated after LIO state estimation is completed. When the number of voxels after the update exceeds the capacity limit, the oldest voxel blocks are evicted in order of their most recent write times.

### C. Cache-Guided Staged KNN Search (C-HKNN)

C-HKNN is used to query neighboring representative points. As shown in Fig. 2, C-HKNN organizes candidate subvoxels around the query point into six search stages according to their spatial distances, containing 27, 54, 44, 54, 72, and 60 subvoxel slots, respectively, for a total of 311 slots spanning at most 60 voxel blocks. The search sequentially visits valid representative points in each stage and maintains the $K$ points closest to the query point, where $K = 5$.

Let $U_g$ be the set of subvoxels in the template that remain unvisited after stage $g$; the lower bound on their distance to the query subvoxel is

$$d_{\text{lb},g} = \min_{u \in \mathcal{U}_g} \text{dist}(\mathcal{B}_q, \mathcal{B}_u) \quad (2)$$

where $\mathcal{B}_q$ and $\mathcal{B}_u$ are the bounding boxes of the corresponding subvoxels. This lower bound is precomputed based on the template and subvoxel resolution. Once $K$ nearest neighbors have been obtained and the distance to the farthest neighbor satisfies $d_K < d_{\text{lb},g}$, the search terminates early; otherwise, it continues through subsequent stages until the entire template has been traversed.

To reduce repeated queries during state iteration, the system maintains an independent cache for each scan point and jointly validates it using the map epoch $e_m$ and the query subvoxel index $\mathbf{f}_i$. The cache stores the addresses of visited blocks and the results of unsuccessful lookups, and records candidate-point addresses by stage. When both the map epoch and query subvoxel index match the cached records, the candidate addresses are reused stage by stage, and distances are recomputed using the current query position to select the nearest neighbors; if further expansion is required, stages not yet cached are searched and added to the records. If either identifier changes, the cache is reset and the search is repeated. The resulting nearest neighbors are used to fit a local plane and construct geometric constraints.

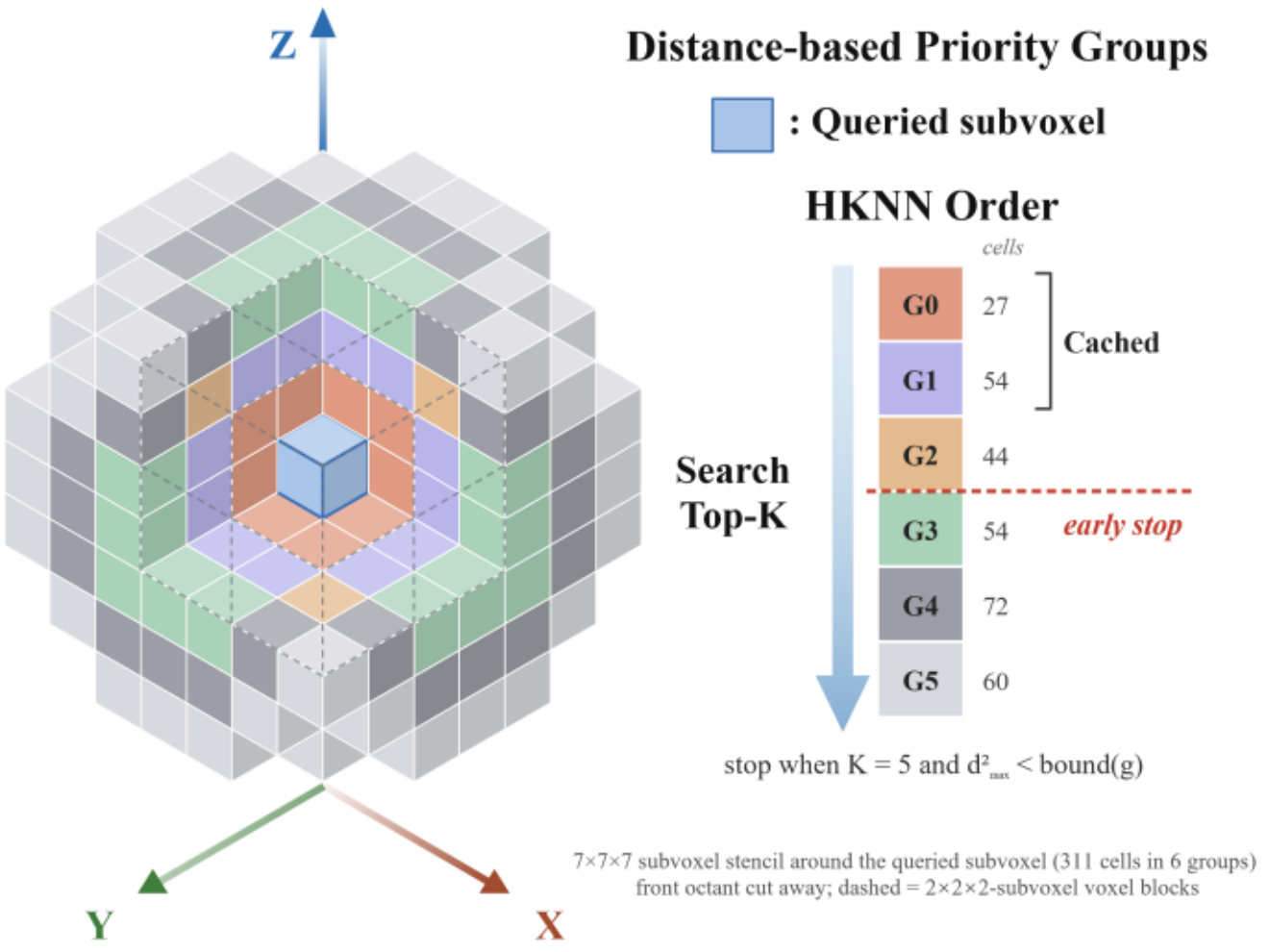

Figure 2. Schematic of the C-HKNN search based on a 7×7×7 sub-voxel template. The candidate sub-voxels surrounding the query sub-voxel (blue) are pre-partitioned into six distance-priority groups (G0-G5, 311 cells in total).

### D. Single Reference Image Patches and Visual Map Updates

The visual map retains only one reference image patch for each map point. During the visual update, the system affinely warps the reference patch of each point using the local plane and relative pose and, after visibility and photometric checks, constructs the pixel residual

$$r_{i,k}^V(\mathbf{x}) = \alpha I\left(\mathbf{u}_{i,k}(\mathbf{x})\right) - \alpha_r I_{r,i,k}^w \quad (3)$$

where $i$ and $k$ denote the map point and the pixel within the patch, respectively, $I_{r,i,k}^w$ is the warped reference pixel intensity, and $\alpha$ and $\alpha_r$ are the current and reference inverse-exposure gains, respectively. Valid residuals are used for coarse-to-fine direct registration in the image pyramid.

The visual map is updated by replacing it as a whole, with each map point and its corresponding reference image patch replaced simultaneously. Let $a$ be the number of frames elapsed since the reference frame was established, $N_{\text{tr}}$ be the current number of valid tracked points, $N_{\text{peak}}$ be the peak number of tracked points since the reference was activated, and $d_{\text{med}}$ be the median displacement between the current projected positions and the reference pixel positions. The conditions for the system to begin generating candidate visual map points are as follows:

$$
\begin{gathered}
a \geq a_{\max}, \\
N_{\mathrm{tr}} < \max(N_{\min}, \lfloor \eta N_{\mathrm{peak}} \rfloor), \\
d_{\mathrm{med}} \geq d_{\max}
\end{gathered} \tag{4}
$$

where $a_{\max}$ limits the duration for which the reference frame is used, $N_{\min}$ and $\eta$ constrain the number of tracked points and the retention ratio relative to its peak, respectively, and $d_{\max}$ limits the pixel displacement.

When any of the above conditions is met, the system generates candidate map points from the current frame. The current frame first undergoes visual correction using the old reference; candidate visual map points are then generated from LiDAR points with valid normals through image-grid, depth, and texture filtering, and one reference image patch from the current frame is extracted for each candidate point. Only when the number of newly generated candidate visual map points in the current frame reaches $N_{\min}$ are the old map points and their patches replaced as a whole and the new reference pose recorded; otherwise, the original visual map is retained. The initial visual map is established using the same procedure.

*E. Reference Pose Uncertainty Maintenance*

Reference image patches are reused over multiple subsequent frames, and pose errors in their source frame continue to affect the photometric observations. Because all patches in the same batch originate from a single frame, the system uses one reference pose $\mathbf{x}_r$ to represent this uncertainty and maintains its joint covariance with the current state $\mathbf{x}$:

$$
\mathbf{P} = \begin{bmatrix} \mathbf{P}_{xx} & \mathbf{P}_{xr} \\ \mathbf{P}_{xr}^T & \mathbf{P}_{rr} \end{bmatrix} \tag{5}
$$

where $\mathbf{P}$ denotes the joint error covariance matrix of the current state $x$ and the reference pose $x_r$. The diagonal blocks $P_{xx}$ and $P_{rr}$ are the error covariance matrices of the current state and the reference pose, respectively. The off-diagonal block $P_{xr}$ represents the cross-covariance between their estimation errors, and $P_{xr}^T$ is its transpose.

The system adopts a Schmidt-type update [10], keeping the mean of the reference pose unchanged. Let the Jacobians of the photometric residual with respect to the current state and reference pose be $\mathbf{H}_x$ and $\mathbf{H}_r$, respectively, and let $\mathbf{H} = [\mathbf{H}_x \mathbf{H}_r]$; the innovation covariance and the current-state gain are

$$
\begin{gathered}
\mathbf{S} = \mathbf{H}\mathbf{P}\mathbf{H}^T + \sigma_I^2 \mathbf{I}, \\
\mathbf{K} = (\mathbf{P}_{xx}\mathbf{H}_x^T + \mathbf{P}_{xr}\mathbf{H}_r^T)\mathbf{S}^{-1}
\end{gathered} \tag{6}
$$

where $\sigma_I^2$ is the pixel observation noise variance. The visual update iteratively corrects the current state using a fixed prior, and the covariance is corrected in a single step after the iterations are complete:

$$
\begin{gathered}
\mathbf{P}_{xx}^+ = \mathbf{P}_{xx} - \mathbf{K}\mathbf{S}\mathbf{K}^T, \\
\mathbf{P}_{xr}^+ = \mathbf{P}_{xr} - \mathbf{K}(\mathbf{H}_x\mathbf{P}_{xr} + \mathbf{H}_r\mathbf{P}_{rr}), \\
\mathbf{P}_{rr}^+ = \mathbf{P}_{rr}
\end{gathered} \tag{7}
$$

The cross-covariance is updated in tandem with IMU propagation and LiDAR correction. After the reference batch is replaced as a whole, the system clones the current posterior pose as the new reference and initializes

$$
\begin{gathered}
\mathbf{P}_{rr}^{\mathrm{new}} = \mathbf{J}\mathbf{P}_{xx}^+\mathbf{J}^T, \\
\mathbf{P}_{xr}^{\mathrm{new}} = \mathbf{P}_{xx}^+\mathbf{J}^T,
\end{gathered} \tag{8}
$$

where $\mathbf{J}$ is the selection matrix for the pose components. When exposure estimation is enabled, the gain and its covariance are first normalized, after which the above initialization is performed so that the new reference retains its correlation with the current state.

**Algorithm 1a: C-HKNN Update**

```
Input: x⁻, P⁻, LiDAR scan 𝒫, map M
x ← x⁻; c ← 0; repeat
│ for p ∈ 𝒫, find up to 5 nearest means via C-HKNN at x
│   fit π_p only with 4–5 means; check λ_min and eigenvalue gaps
│   gate r_p if tests pass and all neighbors are within 0.1 m of π_p
│ if no valid residuals then break
│ stack r,H; solve the fixed-prior iterated EKF
│ if Δx is nonfinite or trial position is > 2 block widths from x⁻ then
│   restore x⁻; restart once with c = 0 and measurement-only gating, or break if already retried
│ apply Δx; count steps below 0.01° and 0.15 mm or the penultimate step if c = 0; correct P and stop at c = 2 or the limit
always insert 𝒫 into M using the resulting x
return x, P, M
```

**Algorithm 1b: Single-Reference-Frame Update**

```
Input: image I_k, state x_k, visual map F
project F in scan-covered voxels; keep the closest per image cell
for q ∈ F(sel) do
│ warp the reference patch; reject invalid patches and photo outliers
│ add valid q to the visual residuals
run the coarse-to-fine EKF with T_ref fixed
apply covariance and cross-covariance reduction
if age ≥ 8 or N_tr < max(N_min,⌊0.6N_peak⌋) or med‖Δu‖ ≥ 12 px then
│ C ← current-frame candidates (|C| ≤ 800)
│ if |C| ≥ 30 then replace reference; clear cache
│ else keep the current reference
on a new reference, normalize exposure gain/covariance if estimated
return x_k and F
```

## IV. Experimental Evaluation

We evaluated the proposed system on both public benchmarks and a custom dataset, and compared it against state-of-the-art methods.

*A. Experimental Setup*

To demonstrate the effectiveness of the proposed efficient LIVO framework, we conducted comprehensive experiments evaluating trajectory accuracy, ablations of key components, computational efficiency, and qualitative mapping performance in diverse environments.

The evaluation was conducted on the Hilti SLAM Challenge 2022 [17] and 2023 [18], NTU VIRAL [19], and Newer College [20] datasets. In addition, we collected an in-house dataset containing multiple sequences using a Livox Mid-360 LiDAR (360° horizontal field of view, 10 Hz) and a HIKROBOT MV-CU013-A0UC camera (1280 × 1024, 10 Hz). The sensor setups were mounted on handheld and unmanned aerial vehicle platforms, respectively (Fig. 3).

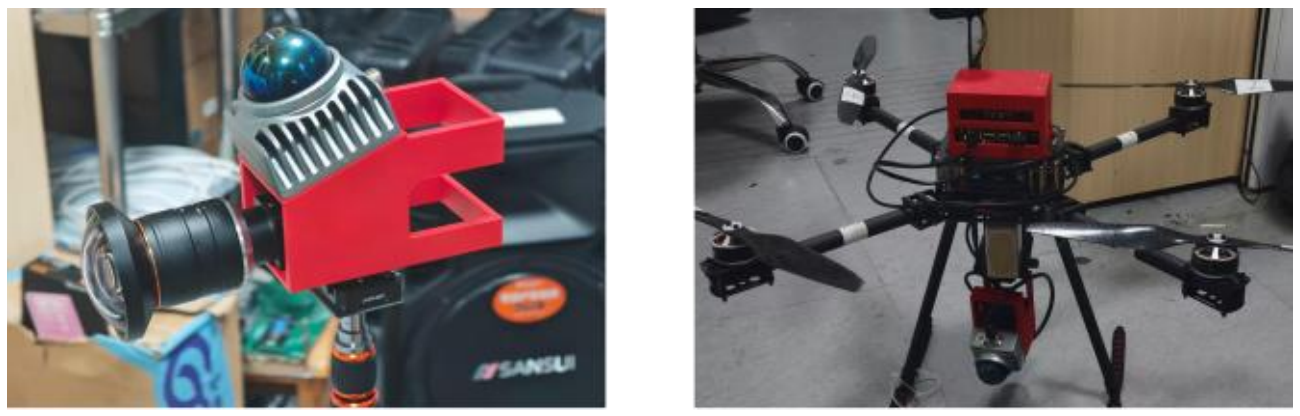

(a) Handheld platform (b) UAV platform

Figure 3. Data-collection platforms used in our custom dataset. (a) Handheld configuration. (b) UAV configuration.

We compared our method with three approaches: FAST-LIVO2, FAST-LIO2, and SR-LIVO [21]. Both the camera and LiDAR data were collected at 10 Hz for evaluation. FAST-LIVO2 and our system used the same LIO and IMU parameters. The ablation study considered three system variants: w/o OctVox, which disables the compact geometric map; w/o SRM, which disables the single reference image patch; and w/o UAF, which disables reference pose uncertainty maintenance. Ours denotes the complete system.

TABLE I. ABSOLUTE TRAJECTORY ERROR ( RMSE IN METERS) ON PUBLIC BENCHMARK DATASETS (X86)

| Dataset Sequence | Sequence | FAST-LIVO2 | FAST-LIO2 | SR-LIVO | Ours (w/o OctVox) | Ours (w/o SRM) | Ours (w/o UAF) | Ours |
|---|---|---|---|---|---|---|---|---|
| **Hilti 2022** | Construction Ground | **0.009** | 0.015 | × | 0.010 | **0.009** | **0.009** | **0.009** |
| | Construction Multilevel | 0.021 | 0.026 | × | 0.021 | **0.020** | **0.020** | **0.020** |
| | Construction Stairs | 0.746 | 4.239 | × | 0.736 | 0.048 | **0.032** | 0.047 |
| | Long Corridor | 0.056 | 0.073 | × | 0.061 | **0.052** | 0.053 | 0.053 |
| | Cupola | **0.272** | × | × | × | 5.191 | 0.670 | 0.530 |
| | Lower Gallery | 0.050 | 19.716 | 3.540 | 25.596 | 0.291 | **0.007** | 0.010 |
| | Attic to Upper Gallery | 1.726 | × | × | **0.300** | 2.218 | 0.584 | 0.526 |
| | Outside Building | 0.045 | 0.174 | 16.441 | 0.043 | **0.028** | 0.048 | 0.047 |
| **Hilti 2023** | Floor 0 | 0.030 | **0.020** | × | 0.032 | 0.028 | 0.028 | 0.028 |
| | Floor 1 | 0.022 | 0.027 | × | 0.016 | **0.015** | **0.015** | **0.015** |
| | Floor 2 | **0.033** | 0.037 | × | 0.040 | 0.037 | 0.039 | 0.038 |
| | Basement | **0.021** | 0.203 | × | 0.024 | 0.023 | 0.023 | 0.022 |
| | Stairs | 0.013 | 0.644 | × | **0.005** | 0.041 | 0.029 | 0.008 |
| | Parking 3x floors down | 0.407 | × | × | **0.276** | 0.323 | 0.382 | 0.462 |
| | Large room | 0.060 | 6.583 | × | 0.049 | 0.035 | 0.881 | **0.025** |
| | Large room(dark) | 0.061 | 5.535 | × | **0.032** | 0.054 | 0.043 | 0.045 |
| **NTU VIRAL** | eee_01 | 0.034 | 0.149 | 0.073 | **0.029** | 0.034 | 0.035 | 0.035 |
| | eee_02 | 0.034 | 0.072 | 0.072 | **0.027** | 0.030 | 0.033 | 0.034 |
| | eee_03 | 0.034 | 0.116 | 0.107 | 0.034 | **0.032** | **0.032** | **0.032** |
| | nya_01 | 0.039 | 0.068 | 0.061 | 0.040 | **0.031** | 0.033 | **0.031** |
| | nya_02 | 0.045 | 0.098 | 0.095 | 0.035 | **0.034** | 0.035 | 0.035 |
| | nya_03 | 0.040 | 0.107 | 0.104 | 0.034 | **0.032** | 0.033 | 0.033 |
| | sbs_01 | **0.033** | 0.090 | 0.088 | 0.034 | **0.033** | 0.034 | **0.033** |
| | sbs_02 | 0.041 | 0.078 | 0.078 | **0.039** | 0.041 | **0.039** | **0.039** |
| | sbs_03 | 0.033 | 0.080 | 0.073 | 0.035 | 0.035 | **0.032** | 0.033 |
| **Newer College** | Quad easy | 0.139 | **0.071** | × | 0.194 | 0.079 | 0.085 | 0.083 |
| | Quad medium | 0.221 | **0.063** | × | 0.241 | 0.079 | 0.077 | 0.077 |
| | Quad hard | 0.125 | 0.083 | × | 0.245 | **0.066** | 0.067 | 0.067 |
| | Stairs | 0.068 | × | × | 0.404 | **0.059** | **0.059** | 0.067 |
| | Underground easy | 0.036 | 0.058 | × | **0.035** | **0.035** | 0.036 | 0.036 |
| | Underground medium | **0.042** | 0.052 | × | 0.045 | **0.042** | **0.042** | **0.042** |
| | Underground hard | **0.049** | 0.066 | × | 0.060 | 0.052 | 0.052 | 0.052 |
| **Average** | | 0.143 | 1.377 | 1.885 | 0.928 | 0.285 | 0.112 | **0.082** |

## B. Accuracy Evaluation

Table I summarizes the absolute trajectory errors (ATE) of all methods on the 32 sequences from four datasets. Bold values indicate the best results, and "×" denotes a failure in trajectory estimation. Instant-LIVO and its Variants achieve the best or tied-best results on 25 of the 32 sequences, spanning construction sites, indoor corridors, underground spaces, and large-scale outdoor environments. Compared with FAST-LIVO2, our method reduces the mean error from 0.143 m to 0.082 m, a reduction of approximately 43%. On the Construction Stairs sequence, the error decreases from 0.746 m to 0.047 m, a reduction of 93.7%.

The ablation study shows that all three modules contribute to the overall performance. Removing OctVox causes the most pronounced degradation: the mean error increases from 0.082 m to 0.928 m, trajectory estimation fails on the Cupola sequence, and the errors on Lower Gallery and Construction Stairs increase to 25.596 m and 0.736 m, respectively. Removing the single-reference-frame visual map (SRM) increases the mean error to 0.285 m, with significant degradation on sequences with challenging visual conditions, such as Cupola (5.191 m vs. 0.530 m), Attic to Upper Gallery (2.218 m vs. 0.526 m), and Lower Gallery (0.291 m vs. 0.010 m). Removing uncertainty-aware fusion (UAF) increases the mean error to 0.112 m, and the error on the Large Room sequence rises from 0.025 m to 0.881 m. Overall, the three modules are complementary, achieving a balance between accuracy and robustness.

## C. Efficiency Analysis

This section evaluates the runtime performance and system-resource usage of all methods on an X86 laptop (i9-12900H, with threads pinned to six performance cores) and an ARM embedded platform (NVIDIA Jetson Orin Nano Super, using all cores). The experiments cover four public datasets and our self-collected dataset, denoted as "We" in the tables.

Table II reports the average per-frame processing time for each dataset on the X86 laptop and ARM embedded platform. For each dataset, we averaged all available sequences to obtain a representative runtime measurement. Across the four public datasets and our own dataset, Instant-LIVO achieved the lowest processing latency on both platforms. It was 2.3× and 2.1× faster than FAST-LIVO2 on the X86 and ARM platforms, respectively, demonstrating a consistent efficiency advantage across different hardware architectures.

Table III quantifies the peak memory usage(RSS) of each dataset on the X86 laptop and ARM embedded platform, which directly affects the ability to perform stable mapping over extended periods. On both platforms, Instant-LIVO required less than one-tenth of the memory used by FAST-LIVO2 while maintaining stable operation. This reduction is attributable to its streamlined geometric map and single visual map-point design, which substantially reduces memory overhead.

Fig. 4 shows the module-level timing distribution in several representative challenging scenarios. Instant-LIVO substantially reduces the time spent in the LIO state-update, LIO map-update, and VIO stages, highlighting the contributions of OctVox, C-HKNN, and the single visual map-point strategy. Fig. 5 reports the memory growth with each processed frame on different platforms for the site-robot-1 sequence from Hilti 2023. Instant-LIVO exhibits a slower memory-growth rate throughout the entire run, owing to its storage of only a single visual map point per geometric map point. This enables stable operation over long sequences under hardware-constrained conditions, such as ARM platforms, and demonstrates a substantially lighter computational load. These figures provide a clearer view of Instant-LIVO's real-time performance.

## D. Qualitative Analysis

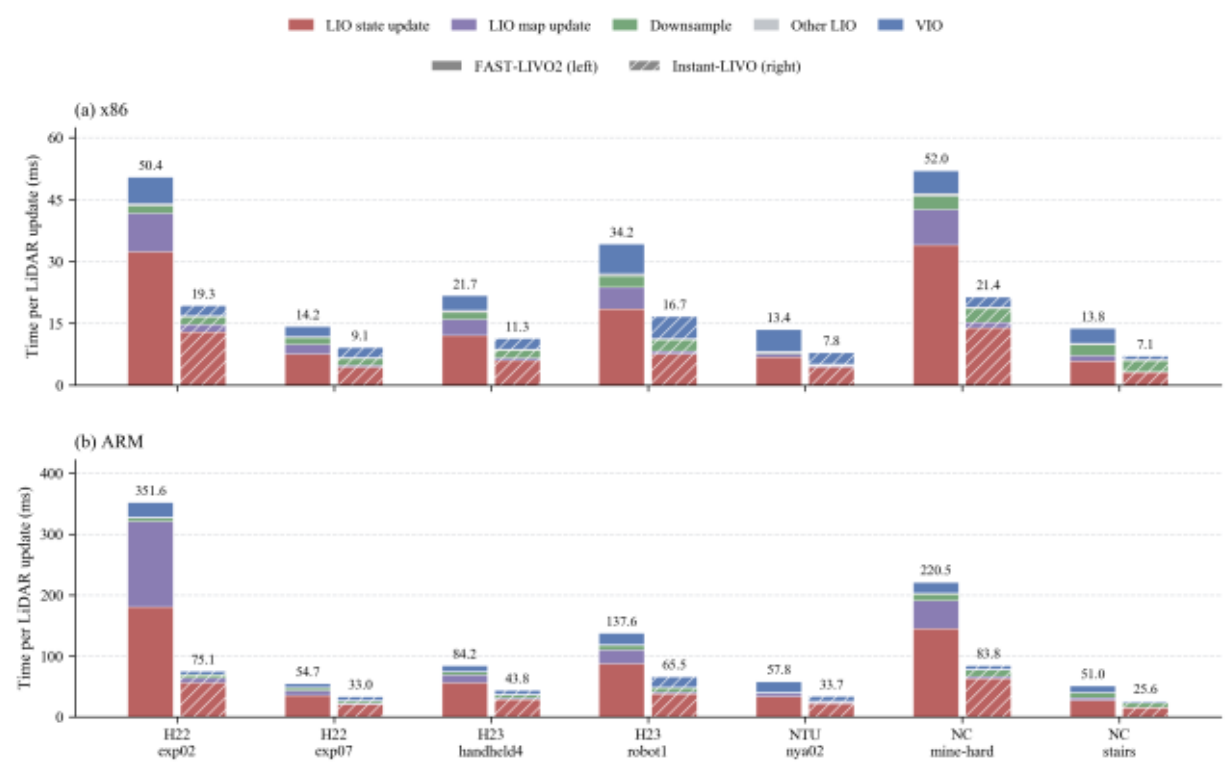

Figure 4. Module-level timing analysis of each state update for FAST-LIVO2 and Instant-LIVO on (a) the X86 platform and (b) the ARM platform. Each bar shows the time spent in the main pipeline stages: LIO state update, LIO map update, downsampling, other LIO operations, and VIO. For each sequence, the methods are ordered from left to right as FAST-LIVO2 (solid) and Instant-LIVO (hatched), with the total time (ms) annotated above each bar.

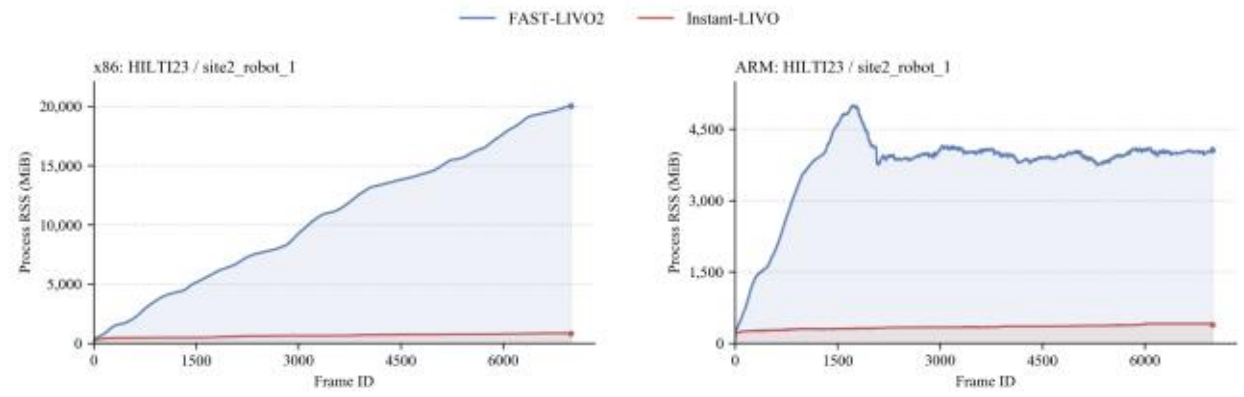

Figure 5. Memory usage as a function of frame index on the site-robot-1 sequence from Hilti 2023.

TABLE II. PER-FRAME RUNTIME(MS), AVERAGED ACROSS DATASETS.

| Dataset Sequence | X86 | | ARM | |
|---|---|---|---|---|
| | FAST-LIVO2 | Ours | FAST-LIVO2 | Ours |
| **Hilti 2022** | 35.45 | 14.91 | 174.04 | 58.48 |
| **Hilti 2023** | 32.83 | 15.47 | 126.98 | 61.67 |
| **NTUVIRAL** | 14.31 | 8.10 | 60.54 | 33.98 |
| **Newer College** | 71.67 | 24.42 | 168.77 | 92.08 |
| **We** | 18.14 | 11.49 | 82.12 | 48.81 |
| **Average** | **34.48** | **14.88** | **122.49** | **59.00** |

TABLE III. AVERAGE PEAK MEMORY USAGE (MIB) ACROSS DATASETS

| Dataset Sequence | X86 | | ARM | |
|---|---|---|---|---|
| | FAST-LIVO2 | Ours | FAST-LIVO2 | Ours |
| **Hilti 2022** | 7247.46 | 375.92 | 4870.57 | 406.45 |
| **Hilti 2023** | 8765.54 | 427.14 | 5178.17 | 296.18 |
| **NTUVIRAL** | 3114.96 | 273.80 | 2989.76 | 204.43 |
| **Newer College** | 15835.36 | 489.66 | 5001.80 | 412.60 |
| **We** | 3701.74 | 347.70 | 3221.25 | 261.68 |
| **Average** | **7733.01** | **382.84** | **4252.31** | **316.27** |

Fig. 6 shows the reconstruction quality of Instant-LIVO on the Hilti SLAM Challenge 2022 dataset. Fig. 7 demonstrates the system's ability to maintain global consistency. Compared with FAST-LIVO2, Instant-LIVO exhibits better alignment in the presence of visible drift. This improvement in consistency results from accounting for uncertainty in the reference pose.

### *E. Discussion*

The experimental results demonstrate that Instant-LIVO achieves superior performance on both public benchmark datasets and challenging self-collected datasets. The OctVox map and single-reference-frame map substantially reduce the system's memory footprint, while C-HKNN accelerates map queries. Accounting for uncertainty in the single reference frame improves system robustness, and the stable performance improvements are further validated on both X86 and ARM platforms.

## V. Conclusion

This letter presented Instant-LIVO, a LiDAR–visual–inertial odometry system designed to balance accuracy, efficiency, and robustness. Its design relies on an OctVox geometric map and a single-reference-frame visual map. Evaluations on public benchmark datasets and custom datasets demonstrate that, compared with state-of-the-art LIVO and LIO methods, the system consistently achieves higher efficiency and lower memory overhead without sacrificing accuracy or robustness. Instant-LIVO is intended to provide a lightweight and fast solution for computation-constrained robotic applications. Future work will explore online coupling with backend optimization.

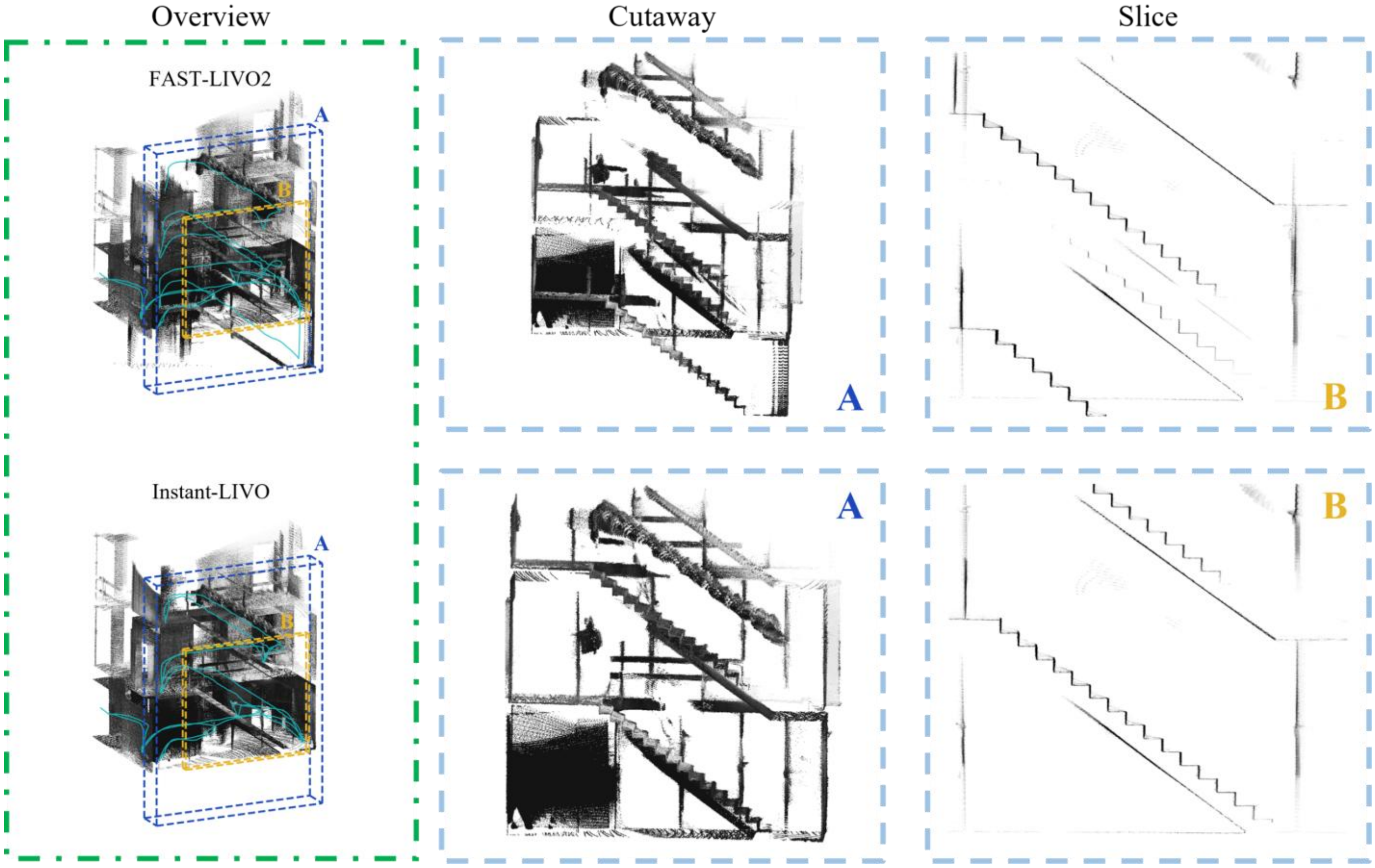

Figure 6. Comparison of reconstruction quality on the Construction Stairs sequence. The green box indicates the overall scene, and the blue box indicates the cropped scene.

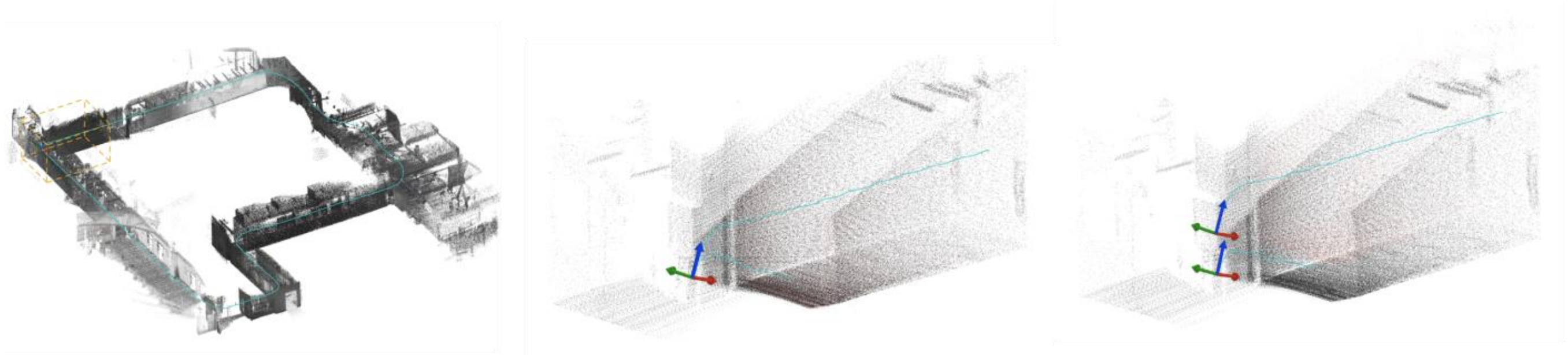

(a) Overall map (b) Instant-LIVO: Start/End alignment (c) FAST-LIVO2: Start/End misalignment

Figure 7. Comparison of start–end alignment consistency on a self-collected corridor sequence. (a) Overall map. (b) Instant-LIVO: accurate alignment between the start and end points. (c) FAST-LIVO2: noticeable drift

## References

[1] C. Campos, R. Elvira, J. J. G. Rodríguez, J. M. M. Montiel, and J. D. Tardós, "ORB-SLAM3: An accurate open-source library for visual, visual-inertial, and multimap SLAM," IEEE Trans. Robot., vol. 37, no. 6, pp. 1874–1890, Dec. 2021, doi: 10.1109/TRO.2021.3075644.

[2] J. Zhang and S. Singh, "LOAM: Lidar odometry and mapping in real-time," in Proc. Robotics: Science and Systems (RSS), Berkeley, CA, USA, Jul. 2014, pp. 1–9, doi: 10.15607/RSS.2014.X.007.

[3] W. Xu, Y. Cai, D. He, J. Lin, and F. Zhang, "FAST-LIO2: Fast direct LiDAR-inertial odometry," IEEE Trans. Robot., vol. 38, no. 4, pp. 2053–2073, Aug. 2022, doi: 10.1109/TRO.2022.3141876.

[4] C. Bai, T. Xiao, Y. Chen, H. Wang, F. Zhang, and X. Gao, "Faster-LIO: Lightweight tightly coupled lidar-inertial odometry using parallel sparse incremental voxels," IEEE Robot. Autom. Lett., vol. 7, no. 2, pp. 4861–4868, Apr. 2022, doi: 10.1109/LRA.2022.3152830.

[5] T. Shan, B. Englot, C. Ratti, and D. Rus, "LVI-SAM: Tightly-coupled lidar-visual-inertial odometry via smoothing and mapping," in Proc. IEEE Int. Conf. Robot. Autom. (ICRA), Xi'an, China, May 2021, pp. 5692–5698, doi: 10.1109/ICRA48506.2021.9561996.

[6] J. Lin and F. Zhang, "R³LIVE: A robust, real-time, RGB-colored, LiDAR-inertial-visual tightly-coupled state estimation and mapping package," in Proc. IEEE Int. Conf. Robot. Autom. (ICRA), Philadelphia, PA, USA, May 2022, pp. 10672–10678, doi: 10.1109/ICRA46639.2022.9811935.

[7] C. Zheng, Q. Zhu, W. Xu, X. Liu, Q. Guo, and F. Zhang, "FAST-LIVO: Fast and tightly-coupled sparse-direct LiDAR-inertial-visual odometry," in Proc. IEEE/RSJ Int. Conf. Intell. Robots Syst. (IROS), Kyoto, Japan, Oct. 2022, pp. 4003–4009, doi: 10.1109/IROS47612.2022.9981107.

[8] C. Zheng, W. Xu, Z. Zou, T. Hua, C. Yuan, D. He, B. Zhou, Z. Liu, J. Lin, F. Zhu, Y. Ren, R. Wang, F. Meng, and F. Zhang, "FAST-LIVO2: Fast, direct LiDAR-inertial-visual odometry," IEEE Trans. Robot., vol. 41, pp. 326–346, 2025, doi: 10.1109/TRO.2024.3502198.

[9] B. Zhou, C. Zheng, Z. Wang, F. Zhu, Y. Cai, and F. Zhang, "FAST-LIVO2 on resource-constrained platforms: LiDAR-inertial-visual odometry with efficient memory and computation," IEEE Robot. Autom. Lett., vol. 10, no. 8, pp. 7931–7938, Aug. 2025, doi: 10.1109/LRA.2025.3581125.

[10] P. Geneva, J. Maley, and G. Huang, "An efficient Schmidt-EKF for 3D visual-inertial SLAM," in Proc. IEEE/CVF Conf. Comput. Vis. Pattern Recognit. (CVPR), Long Beach, CA, USA, 2019, pp. 12097–12107, doi: 10.1109/CVPR.2019.01238.

[11] J. Lin, C. Zheng, W. Xu, and F. Zhang, "R²LIVE: A robust, real-time, LiDAR-inertial-visual tightly-coupled state estimator and mapping," IEEE Robot. Autom. Lett., vol. 6, no. 4, pp. 7469–7476, Oct. 2021, doi: 10.1109/LRA.2021.3095515.

[12] S. Hong, C. Zheng, Y. Shen, C. Li, F. Zhang, T. Qin, and S. Shen, "GS-LIVO: Real-time LiDAR, inertial, and visual multisensor fused odometry with Gaussian mapping," IEEE Trans. Robot., vol. 41, pp. 4253–4268, 2025, doi: 10.1109/TRO.2025.3582809.

[13] S. Zhou, Z. Wang, X. Dai, W. Song, and S. Gu, "LIR-LIVO: A lightweight, robust LiDAR/vision/inertial odometry with illumination-resilient deep features," 2025, arXiv:2502.08676. [Online]. Available: https://arxiv.org/abs/2502.08676

[14] H. Zhang, J. Yuan, Z. Hu, L. Du, S. Bao, W. Wan, and K. Harada, "Voxel-LIVO: Accurate and robust LiDAR-inertial-visual fused odometry and mapping in challenge environment," IEEE Trans. Instrum. Meas., vol. 75, Art. no. 9505317, 2026, doi: 10.1109/TIM.2026.3657488.

[15] T. Shan, B. Englot, D. Meyers, W. Wang, C. Ratti, and D. Rus, "LIO-SAM: Tightly-coupled lidar inertial odometry via smoothing and mapping," in Proc. IEEE/RSJ Int. Conf. Intell. Robots Syst. (IROS), Las Vegas, NV, USA, Oct. 2020, pp. 5135–5142, doi: 10.1109/IROS45743.2020.9341176.

[16] L. Wang, X. Zhang, C. Li, D. He, Y. Pan, and J. Yi, "Super-LIO: A robust and efficient LiDAR-inertial odometry system with a compact mapping strategy," IEEE Robot. Autom. Lett., vol. 11, no. 3, pp. 2666–2673, Mar. 2026, doi: 10.1109/LRA.2026.3653372.

[17] L. Zhang, M. Helmberger, L. F. T. Fu, D. Wisth, M. Camurri, D. Scaramuzza, and M. Fallon, "Hilti-Oxford dataset: A millimeter-accurate benchmark for simultaneous localization and mapping," IEEE Robot. Autom. Lett., vol. 8, no. 1, pp. 408–415, Jan. 2023, doi: 10.1109/LRA.2022.3226077.

[18] A. D. Nair, J. Kindle, P. Levchev, and D. Scaramuzza, "Hilti SLAM Challenge 2023: Benchmarking single + multi-session SLAM across sensor constellations in construction," IEEE Robot. Autom. Lett., vol. 9, no. 8, pp. 7286–7293, Aug. 2024, doi: 10.1109/LRA.2024.3421791.

[19] T.-M. Nguyen, S. Yuan, M. Cao, Y. Lyu, T. H. Nguyen, and L. Xie, "NTU VIRAL: A visual-inertial-ranging-lidar dataset, from an aerial vehicle viewpoint," Int. J. Robot. Res., vol. 41, no. 3, pp. 270–280, Mar. 2022, doi: 10.1177/02783649211052312.

[20] M. Ramezani, Y. Wang, M. Camurri, D. Wisth, M. Mattamala, and M. Fallon, "The Newer College dataset: Handheld LiDAR, inertial and vision with ground truth," in Proc. IEEE/RSJ Int. Conf. Intell. Robots Syst. (IROS), Las Vegas, NV, USA, Oct. 2020, pp. 4353–4360, doi: 10.1109/IROS45743.2020.9340849.

[21] Z. Yuan, J. Deng, R. Ming, F. Lang, and X. Yang, "SR-LIVO: LiDAR-inertial-visual odometry and mapping with sweep reconstruction," IEEE Robot. Autom. Lett., vol. 9, no. 6, pp. 5110–5117, Jun. 2024, doi: 10.1109/LRA.2024.3389415.